\documentclass[11pt,a4paper]{article}
\usepackage[hyperref]{acl2019}
\usepackage{times}
\usepackage{latexsym}

\usepackage{url}

\usepackage[ngerman]{babel} 
\usepackage[utf8]{inputenc}

\usepackage{xcolor}
\usepackage{dirtytalk}

\usepackage{graphicx}
\usepackage{multirow}
\usepackage{placeins}
\usepackage{arydshln}
\usepackage{amsmath}

\usepackage{float}

\aclfinalcopy 

\title{Selecting Artificially-Generated Sentences for Fine-Tuning Neural Machine Translation}
\author{Alberto Poncelas \and Andy Way\\
ADAPT Centre, School of Computing, \\ Dublin City University, Dublin, Ireland\\ {\tt \{firstname.lastname\}@adaptcentre.ie}
}

\date{}

\begin{document}
\maketitle
\begin{abstract}


Neural Machine Translation (NMT) models tend to achieve best performance when larger sets of parallel sentences are provided for training. For this reason, augmenting the training set with artificially-generated sentence pairs can boost performance.

Nonetheless, the performance can also be improved with a small number of sentences if they are in the same domain as the test set. Accordingly, we want to explore the use of artificially-generated sentences along with data-selection algorithms to improve German-to-English NMT models trained solely with authentic data.

In this work, we show how artificially-generated sentences can be more beneficial than authentic pairs, and demonstrate their advantages when used in combination with data-selection algorithms.


%




\end{abstract}

\section{Introduction}

The data used for training Machine Translation (MT) models consist mainly of a set of parallel sentences (a set of sentence-pairs in which each sentence is paired with its translation). As Neural Machine Translation (NMT) models typically achieve best performance when using large sets of parallel sentences, they can benefit from the sentences created by Natural Language Generation (NLG) systems. Although artificial data is expected to be of lower quality than authentic sentences, it still can help the model to learn how to better generalize over the training instances and produce better translations. 

A popular technique used to create artificial data is the back-translation technique \citep{sennrich2015improving,poncelas2018investigating}. This consists of generating sentences in the source language by translating monolingual sentences in the target language. Then, these sentences in both languages are paired and can be used to augment the original parallel training set used to build better NMT models.




Nonetheless, if synthetic data are not in the same domain as the test set, it can also hurt the performance. For this reason,  we explore an alternative approach to better use the artificially-generated training instances to improve NMT models. In particular, we propose that instead of blindly adding back-translated sentences into the training set they can be considered as candidate sentences for a data-selection algorithm to decide which sentence-pairs should be used to fine-tune the NMT model. By doing that, instead of increasing the number of training instances in a motivated manner, the generated sentences provide us with more chances of obtaining relevant parallel sentences (and still use smaller sets for fine-tuning).

As we want to build task-specific NMT models, in this work we explore two data-selection algorithms that are classified as Transductive Algorithms (TA): Infrequent {\em N}-gram Recovery (INR) and Feature Decay Algorithms (FDA). These methods use the test set $S_{test}$ (the document to be translated) as the seed to retrieve sentences. In transductive learning \citep{Vapnik1998} the goal is to identify the best training instances to learn how to classify a given test set. In order to select these sentences, the TAs search for those {\em n}-grams in the test set that are also present in the source side of the candidate sentences.






Although augmenting the candidate pool with more sentences should be beneficial, as the TAs select the sentences based on overlapping {\em n}-gram the mistakes produced by the model used for back-translation (which are those commonly addressed in NLG such as the generated word order or word choice) can be a disadvantage.

In this work, we explore whether TAs are more inclined to select authentic or artificial sentences. In addition, we propose three different methods of how they can be combined into a single hybrid set. Finally, we investigate whether the hybrid sets retrieved by TAs can be more useful than the authentic set of sentences to fine-tune NMT models.

\section{Related Work}

The work presented in this paper is based on two main concepts: the generation of synthetic sentences, and the selection of sentences from a set $S$ of candidates.

\subsection{Use of Artificially-Generated Data to Improve MT Models}
\label{sec:synthetic_data}

The proposal of \citet{sennrich2015improving} showed that NMT models can be improved by back-translating a set of (monolingual) sentences in the target side into the source side using an MT model. Other uses of monolingual target-side sentences include building the parallel set by using a NULL token in the source side \citep{sennrich2015improving} or creating language models to improve the decoder \citep{Glehre2015}.




\citet{hoang2018iterative} improve the model used for back-translation by training this model with increasing amounts of artificial sentences. They iteratively improve the models creating artificial sentences of better quality.


Similarly to this paper, the use of artificially-generated sentences to fine-tuned models has also been explored by \citet{chinea2017adapting} where they select monolingual authentic sentences in the source-side and translate them into the target language, or the work of \citet{poncelas2019adaptation} where they use back-translated sentences only to adapt the models.



\subsection{Adaptation of NMT Models to the Test Set}

The improvement of NMT models can be performed by fine-tuning \citep{luong2015stanford,freitag2016fast}, i.e. train the models for additional epochs using a small set of in-domain data. Alternatively, \citet{van2017dynamic} train models using smaller but more in-domain sentences in each epoch of the training process.

The use of the test set to retrieve relevant sentences for fine-tuning the model has been explored by \citet{li2016one}, adapting a different model for each sentence in the test set, or \citet{poncelas2018feature,poncelas2019transductive} where they adapt the model for the complete test set using transductive data-selection algorithms.


\section{Transductive Algorithms}
\label{sec:Transductive_algorithms}

In this paper, the sentences used to fine-tune the model are retrieved using INR and FDA. These methods select sentences by scoring each sentence $s$ from the candidate pool $S$, and adding that with the highest score to a selected pool $L$. This process is performed iteratively until the selected pool contains $N$ sentences.

\paragraph{Infrequent {\em n}-gram Recovery} (INR) \citep{parcheta2018data,gasco2012does}: This method selects those sentences that contain {\em n}-grams from the test set that are infrequent (ignoring frequent words such as stop words or general-domain terms). A candidate sentence $s \in S$ is scored according to the number of infrequent {\em n}-grams shared with the set of sentences of the test set $S_{test}$, computed as in \eqref{eq:infreq_ngr_recover}:

\begin{equation}\label{eq:infreq_ngr_recover}
score(s)=\sum_{ngr \in \{ S_{test} \bigcap  s \} }  max(0,t-C_L(ngr)) 
\end{equation}

\noindent where $t$ is the threshold that indicates the number of occurrences of an {\em n}-gram to be considered infrequent. If the number of occurrences of $ngr$ in the selected pool ($C_L(ngr)$) is above the threshold $t$, then the component $max(0,t-C_S(ngr))$ is 0 and so the {\em n}-gram does not contribute to scoring the sentence.


%
\paragraph{Feature Decay Algorithms} \citep{biccici2011instance,biccici2013feature} also retrieve those sentences sharing the highest number of {\em n}-grams from the test set. However, in order to increase the variability and avoid selecting the same {\em n}-grams, those that have been selected are penalized is proportional to the number of occurrences in $L$. The score of a sentence is computed as in \eqref{eq:fda_sentencescore}:


\begin{equation}\label{eq:fda_sentencescore}
score(s,L)=\frac{\sum_{ngr \in S_{test}} 0.5^{C_L(ngr)}}{length(s)}
\end{equation}

\noindent where $length(s)$ indicates the number of words in the sentence $s$. According to the equation, the more occurrences of $ngr$ in $L$, the smaller the contribution is to the scoring of the sentence $s$. 



\subsection{Models Adapted with Hybrid Data}
\label{ssec:hybrid_data_creation}


In order to fine-tune models with hybrid data, we propose three methods of creating these sets: \textit{hybr}, \textit{batch} and \textit{online}. These methods can be classified depending on whether the combination is performed before or after the execution of the TA.

\paragraph{Combine Before Selection.} This approach consists of selecting from a hybrid set (\textit{hybr}). This involves concatenating both the authentic candidate $S_{auth}$ and artificial $S_{synth}$ sentences as a first step and then executing the TAs with the new candidate set $S_{auth+synth}$. 




\paragraph{Combine After Selection. } Another approach is to force the presence of both authentic and synthetic sentences by using different proportions of TA-selected authentic ($L_{auth}$) and synthetic ($L_{synth}$) sentence pairs. We concatenate the top-$(N*\gamma)$ sentences from the selected authentic set and the top-$(N*(1-\gamma))$ from the synthetic set. The value of $\gamma \in [0,1]$ indicates the proportion of authentic and synthetic sentences. For example, $\gamma=0.75$ indicates that the 75\% of sentences in the dataset are authentic and the remaining 25\% are artificially generated.

The selected synthetic set $L_{synth}$ can be obtained by executing the TAs on artificial candidate sentences $S_{synth}$ (\textit{batch}). This implies that the sentences will be retrieved by finding overlaps of {\em n}-grams between the test set and artificial sentences. 

Alternatively, the retrieval may be carried by finding overlaps in the target-side (\textit{online}) as they are human-produced sentences. However, as the test set is in the source language, we need to first generate an approximated translation of the test with a general-domain MT model \citep{poncelas2018data,poncelas2018adapt}. Unlike in \textit{batch}, the advantage of this approach is that it is not necessary to generate the source side of the whole set of monolingual sentences, but rather only those selected by the TA.

\section{Experiments}

\subsection{Data and Models Settings}
\label{sec:data_model_settings}

We build German-to-English NMT models using the following datasets:

\begin{itemize}
    \item Training data: German-English parallel sentences provided in WMT 2015 \citep{bojar-EtAl:2015:WMT} (4.5M sentence pairs).
    \item Test sets: We evaluate the models with two test sets in different domains:
        \begin{itemize}
        \item \textit{BIO test set}: the Cochrane\footnote{\url{http://www.himl.eu/test-sets}} dataset from the WMT 2017 biomedical translation shared task \citep{yepes2017findings}.
        \item \textit{NEWS test set}: The test set provided in WMT 2015 News Translation Task.
        \end{itemize}
\end{itemize}

All these data sets are tokenized, truecased, and Byte Pair Encoding (BPE) \citep{sennrich2016neural} is applied using 89,500 merge operations.


The NMT models are built using the attentional encoder-decoder framework with OpenNMT-py\footnote{\url{https://github.com/OpenNMT/OpenNMT-py}} \citep{opennmt}. We use the default values in the parameters: 2-layer LSTM \citep{hochreiter1997long} with 500 hidden units. The size of the vocabulary is 50,000 words for each language.

In order to retrieve sentences, we use the TAs with default configuration (using {\em n}-grams of order 3 to find overlaps between the seed and the training data) to extract sets of 100K, 200K, and 500K sentences. We use a threshold of $t=40$ for INR although this causes the INR to retrieve less than 500K sentences. Accordingly, the results shown for INR will include only 100K and 200K sentences.

\subsection{Back-Translation Generation Settings}

In order to generate artificial sentences, we use an NMT model (we refer to it as \textit{BT model}) to back-translate sentences from the target language into the source language. This model is built by training a model with 1M sentences sampled from the training data and using the same configuration described above (but in the reverse language direction, English-to-German).

As we want to compare authentic and synthetic sentences, we back-translate the target-side of the training data using the BT model. By doing this we ensure both sets are comparable which allows us to perform a fair analysis of whether artificial sentences are more likely to be selected by a TA and which are more useful to fine-tune the models.

Note also that there are 1M sentences that have been generated by translating the same target-side sentences used in training. This could cause the generated sentences to be exactly the same as authentic ones. However, this is not always the case as we report in Section~\ref{ssec:analysis_back_translated_sentences}.








\section{Results}

\begin{table}[!htbp]
\centering
\begin{center}
\begin{tabular}{ |p{0.2cm}|p{1.2cm}|p{1.1cm}|p{1.1cm}|p{1.1cm}|p{1.1cm}|}
\hline
&& \footnotesize BASE13	&	\footnotesize BASE12 + INR	&	\footnotesize BASE12 + FDA\\
\hline
\multicolumn{5}{|c|}{ BIO } \\
\hline
\multirow{3}{*}{\rotatebox[origin=c]{90}{\centering  100K lines}}
&BLEU	&	33.14	&	\bf33.52*	&	\bf33.68*	\\
&TER 	&	46.79	&	\bf45.92*	&	\bf45.97*	\\
&MET.	&	34.57	&	\bf34.77	&	\bf34.71	\\
&CHRF3	&	59.08	&	\bf59.43	&	\bf59.24	\\
\hline							
\multirow{3}{*}{\rotatebox[origin=c]{90}{\centering  200K lines}}							
&BLEU	&	33.14	&	\bf33.88*	&	\bf33.96*	\\
&TER 	&	46.79	&	\bf45.90*	&	\bf45.64*	\\
&MET.	&	34.57	&	\bf34.94*	&	\bf35.01*	\\
&CHRF3	&	59.08	&	\bf59.56	&	\bf59.56	\\
\hline							
\multirow{3}{*}{\rotatebox[origin=c]{90}{\centering  500K lines}}							
&BLEU	&	33.14	&	-	&	\bf33.75*	\\
&TER 	&	46.79	&	-	&	\bf45.92*	\\
&MET.	&	34.57	&	-	&	\bf34.92*	\\
&CHRF3	&	59.08	&	-	&	\bf59.57	\\
\hline	
\multicolumn{5}{|c|}{ NEWS } \\							
\hline							
\multirow{3}{*}{\rotatebox[origin=c]{90}{\centering  100K lines}}							
&BLEU	&	26.34	&	\bf26.49	&	\bf26.49	\\
&TER 	&	54.41	&	\bf54.19	&	\bf54.21	\\
&MET.	&	30.09	&	\bf30.21*	&	\bf30.21*	\\
&CHRF3	&	51.71	&	\bf51.78	&	\bf51.80	\\
\hline							
\multirow{3}{*}{\rotatebox[origin=c]{90}{\centering  200K lines}}							
&BLEU	&	26.34	&	\bf26.44	&	\bf26.55*	\\
&TER 	&	54.41	&	\bf54.35	&	\bf54.17*	\\
&MET.	&	30.09	&	\bf30.12	&	\bf30.24*	\\
&CHRF3	&	51.71	&	51.67	&	\bf51.89	\\
\hline							
\multirow{3}{*}{\rotatebox[origin=c]{90}{\centering  500K lines}}							
&BLEU	&	26.34	&	-	&	\bf26.40*	\\
&TER 	&	54.41	&	-	&	54.47	\\
&MET.	&	30.09	&	-	&	\bf30.10*	\\
&CHRF3	&	51.71	&	-	&	\bf51.71	\\
\hline					
\end{tabular}
\caption{ 
Performance of the BASE13 model, and the models fine-tuned with subsets of the training data.
\label{table:baselines}
 }
\end{center}
\end{table}

First of all, we present in Table~\ref{table:baselines} the performance of the model trained with all data for 13 epochs (BASE13), as this is when the model converges. We also show the performance of the model when fine-tuning the 12th epoch with the subset of (authentic) data selected by INR (\textit{INR} column) and FDA (\textit{FDA} column).

In order to evaluate the performance of the models, we present the following evaluation metrics: BLEU \citep{papineni2002bleu}, TER \citep{snover2006study}, METEOR \citep{banerjee2005meteor}, and CHRF \citep{popovic2015chrf}. These metrics provide an estimation of the translation quality when the output is compared to a human-translated reference. Note that in general, the higher the score, the better the translation quality is. The only exception is TER which is an error metric and so lower results indicate better quality.

In addition, we indicate in bold those scores that show an improvement over the baseline (in Table~\ref{table:baselines} we use BASE13 as the baseline) and add an asterisk if the improvements are statistically significant at p=0.01 (using Bootstrap Resampling \citep{koehn04}, computed with multeval \citep{clark2011better}).

In the table, we can see that using a small subset of data for training the 13th epoch can cause the performance of the model to improve. In the following experiments, we want to compare whether augmenting the candidate set with synthetic data can further boost these improvements. For this reason, we use \textit{INR} and FDA \textit{FDA} as baselines.

\subsection{Results of Models Fine-tuned with Hybrid Data}

\begin{table}[!ht] 
\centering
\begin{center}
\begin{tabular}{ |p{0.2cm}|p{1.5cm}|p{1.5cm}|p{1.5cm}|}
\hline
&&	INR	&	INR HYBR	\\
\hline		
\multicolumn{4}{|c|}{BIO}\\					
\hline
\multirow{4}{*}{\rotatebox[origin=c]{90}{\centering  100K lines}}
&BLEU	&	33.52	&	\bf33.87	\\
&TER	&	45.92	&	46.17	\\
&METEOR	&	34.77	&	\bf35.01	\\
&CHRF3	&	59.43	&	\bf59.53	\\
\hline					
\multirow{4}{*}{\rotatebox[origin=c]{90}{\centering 200K lines}}				
&BLEU	&	33.88	&	33.70	\\
&TER	&	45.90	&	46.33	\\
&METEOR	&	34.94	&	\bf35.23	\\
&CHRF3	&	59.56	&	\bf60.03	\\
\hline					
\multicolumn{4}{|c|}{NEWS}\\				
\hline					
\multirow{4}{*}{\rotatebox[origin=c]{90}{\centering  100K lines}}	&BLEU	&	26.49	&	\bf26.76	\\
&TER	&	54.19	&	54.36	\\
&METEOR	&	30.21	&	\bf30.48*	\\
&CHRF3	&	51.78	&	\bf52.35*	\\
\hline					
\multirow{4}{*}{\rotatebox[origin=c]{90}{\centering  200K lines}}	&BLEU	&	26.44	&	\bf26.80*	\\
&TER	&	54.35	&	\bf54.34	\\
&METEOR	&	30.12	&	\bf30.51*	\\
&CHRF3	&	51.67	&	\bf52.39*	\\
\hline	
\end{tabular}
\caption{ 
Results of the models built with different sizes of INR-selected hybrid data following the \textit{hybr} approach. The results in bold indicate an improvement over INR. The asterisk means the improvement is statistically significant at p=0.01.
\label{table:results_INR_hybr}}
\end{center}
\end{table}

\begin{table}[!ht] 
\centering
\begin{center}
\begin{tabular}{ |p{0.2cm}|p{1.5cm}|p{1.5cm}|p{1.5cm}|}
\hline
&&	FDA	&	FDA HYBR	\\
\hline		
\multicolumn{4}{|c|}{BIO}\\					
\hline
\multirow{4}{*}{\rotatebox[origin=c]{90}{\centering  100K lines}}
&BLEU	&	33.68	&	\bf33.86	\\
&TER	&	45.97	&	46.20	\\
&METEOR	&	34.71	&	\bf35.22*	\\
&CHRF3	&	59.24	&	\bf59.89*	\\
\hline					
\multirow{4}{*}{\rotatebox[origin=c]{90}{\centering	200K lines}}				
&BLEU	&	33.96	&	33.94	\\
&TER	&	45.64	&	46.09	\\
&METEOR	&	35.01	&	\bf35.29	\\
&CHRF3	&	59.56	&	\bf59.89	\\
\hline					
\multirow{4}{*}{\rotatebox[origin=c]{90}{\centering  500K lines}}					
&BLEU	&	33.75	&	\bf33.90	\\
&TER	&	45.92	&	46.34	\\
&METEOR	&	34.92	&	\bf35.12	\\
&CHRF3	&	59.57	&	\bf59.80	\\
\hline					
\multicolumn{4}{|c|}{NEWS}\\				
\hline					
\multirow{4}{*}{\rotatebox[origin=c]{90}{\centering  100K lines}}
&BLEU	&	26.49	&	\bf26.71	\\
&TER	&	54.21	&	54.31	\\
&METEOR	&	30.21	&	\bf30.43*	\\
&CHRF3	&	51.80	&	\bf52.30*	\\
\hline					
\multirow{4}{*}{\rotatebox[origin=c]{90}{\centering	200K lines}}				
&BLEU	&	26.55	&	\bf26.78*	\\
&TER	&	54.17	&	54.42	\\
&METEOR	&	30.24	&	\bf30.51*	\\
&CHRF3	&	51.89	&	\bf52.41*	\\
\hline					
\multirow{4}{*}{\rotatebox[origin=c]{90}{\centering	500K lines}}				
&BLEU	&	26.40	&	\bf26.78*	\\
&TER	&	54.47	&	\bf54.38	\\
&METEOR	&	30.10	&	\bf30.51*	\\
&CHRF3	&	51.71	&	\bf52.38*	\\
\hline	
\end{tabular}
\caption{ 
Results of the models built with different sizes of FDA-selected hybrid data following the \textit{hybr} approach. The results in bold indicate an improvement over FDA. The asterisk means the improvement is statistically significant at p=0.01.
\label{table:results_FDA_hybr}}
\end{center}
\end{table}

In the first set of experiments we explore the \textit{hybr} approach, i.e. the TAs are executed on a mixture of authentic and synthetic data (combined before the execution of the TA). We present the results of the models trained with these sets in Table~\ref{table:results_INR_hybr} (for INR) and Table~\ref{table:results_FDA_hybr} (for FDA). In the first column, we include as the baseline the fine-tuned models presented in Table \ref{table:baselines}.

The results in the tables show that increasing the size of the candidate pool is beneficial. We see that most scores are better (marked in bold) than the model fine-tuned with only authentic data. However, the performance is also dependent on the domain. When comparing BIO and NEWS subtables we see that the models adapted for the latter domain tend to achieve better performances as most of the scores are statistically significant improvements.


When analyzing the selected dataset we find that the authentic sentences constitute slightly above half (between 51\% and 64\% of the sentences). This is an indicator that artificially-generated sentences contain {\em n}-grams that can be found by TA and are as useful as authentic sentences.

In addition, the amount of duplicated target-side sentences is very low (between 10\% and 13\%). This indicates that the MT-generated sentences contain {\em n}-grams that are different from the authentic counterpart, which increases the variety of the candidates that are useful for the TA to select.




\begin{table*}[!ht] 
\centering
\begin{center}
\begin{tabular}{ |p{0.2cm}|p{2cm}||p{1.2cm}||p{1.2cm}|p{1.2cm}|p{1.2cm}||p{1.2cm}|p{1.2cm}|p{1.2cm}|}
\hline
\multicolumn{3}{|c|}{}& \multicolumn{3}{|c|}{batch} &\multicolumn{3}{|c|}{online} \\
\hline
&&	INR	& $\gamma=0.75$ & $\gamma=0.50$ & $\gamma=0.25$ & $\gamma=0.75$ & $\gamma=0.50$ & $\gamma=0.25$ \\
\hline
\multicolumn{9}{|c|}{BIO}\\
\hline
\multirow{4}{*}{\rotatebox[origin=c]{90}{\centering 100K lines}}
&	BLEU	&	33.52	&	\bf33.88	&	33.50	&	\bf33.67	&	33.46	&	33.62	&	33.13	\\
&	TER	&	45.92	&	46.23	&	46.63	&	46.60	&	46.40	&	46.39	&	46.64	\\
&	METEOR	&	34.77	&	\bf35.03	&	\bf35.13	&	\bf35.13	&	\bf34.90	&	\bf34.96	&	\bf35.12	\\
&	CHRF3	&	59.43	&	\bf59.62	&	\bf59.72	&	\bf59.95	&	\bf59.48	&	\bf59.73	&	59.92	\\
\hline																	
\multirow{4}{*}{\rotatebox[origin=c]{90}{\centering 200K lines}}																	
&	BLEU	&	33.88	&	33.85	&	33.70	&	33.51	&	33.69	&	33.52	&	33.39	\\
&	TER	&	45.90	&	46.43	&	46.14	&	46.80	&	46.40	&	46.46	&	46.62	\\
&	METEOR	&	34.94	&	\bf35.00	&	\bf35.20	&	\bf35.06	&	\bf35.04	&	\bf34.96	&	\bf35.08	\\
&	CHRF3	&	59.56	&	59.51	&	\bf59.95	&	\bf59.93	&	\bf59.61	&	59.49	&	\bf59.91	\\
\hline																	
\multicolumn{9}{|c|}{NEWS}\\																	
\hline																	
\multirow{4}{*}{\rotatebox[origin=c]{90}{\centering  100K lines}}																	
&	BLEU	&	26.49	&	\bf26.81*	&	\bf26.59	&	\bf26.75	&	\bf26.77*	&	\bf26.73	&	\bf26.75	\\
&	TER	&	54.19	&	54.39	&	54.46	&	54.49	&	54.4	&	54.71	&	54.84	\\
&	METEOR	&	30.21	&	\bf30.45*	&	\bf30.51*	&	\bf30.65*	&	\bf30.54*	&	\bf30.51*	&	\bf30.64*	\\
&	CHRF3	&	51.78	&	\bf52.30*	&	\bf52.46*	&	\bf52.69*	&	\bf52.34*	&	\bf52.46*	&	\bf52.70*	\\
\hline																	
\multirow{4}{*}{\rotatebox[origin=c]{90}{\centering  200K lines}}								
&	BLEU	&	26.44	&	\bf26.85*	&	\bf26.77*	&	\bf26.61	&	\bf26.79*	&	\bf26.81*	&	\bf26.66	\\
&	TER	&	54.35	&	54.37	&	54.43	&	54.70	&	54.4	&	54.67	&	54.67\\	
&	METEOR	&	30.12	&	\bf30.49*	&	\bf30.58*	&	\bf30.60*	&	\bf30.48*	&	\bf30.55*	&	\bf30.63*	\\
&	CHRF3	&	51.67	&	\bf52.35*	&	\bf52.55*	&	\bf52.60*	&	\bf52.25*	&	\bf52.57*	&	\bf52.69*	\\

\hline
\end{tabular}
\caption{Results of the models built with different sizes of INR-selected hybrid data following the \textit{batch} and  \textit{online} approaches. The results in bold indicate an improvement over INR. The asterisk means the improvement is statistically significant at p=0.01. \label{table:results_INR_batch_online}}
\end{center}
\end{table*}

\begin{table*}[!ht] 
\centering
\begin{center}
\begin{tabular}{ |p{0.2cm}|p{2cm}||p{1.2cm}||p{1.2cm}|p{1.2cm}|p{1.2cm}||p{1.2cm}|p{1.2cm}|p{1.2cm}|}
\hline
\multicolumn{3}{|c|}{}& \multicolumn{3}{|c|}{batch} &\multicolumn{3}{|c|}{online} \\
\hline
&&	FDA	& $\gamma=0.75$ & $\gamma=0.50$ & $\gamma=0.25$ & $\gamma=0.75$ & $\gamma=0.50$ & $\gamma=0.25$ \\
\hline
\multicolumn{9}{|c|}{BIO}\\
\hline
\multirow{4}{*}{\rotatebox[origin=c]{90}{\centering 100K lines}}
&BLEU	&	33.68	&	33.56	&	33.48	&	33.3	&	33.39	&	33.41	&	33.45	\\
&TER	&	45.97	&	46.51	&	46.49	&	46.93	&	46.33	&	46.58	&	46.7	\\
&METEOR	&	34.71	&	\bf34.97	&	\bf35.09*	&	\bf34.89	&	\bf34.97	&	\bf35.00	&	\bf35.11*	\\
&CHRF3	&	59.24	&	\bf59.48	&	\bf59.62	&	\bf59.59	&	\bf59.45	&	\bf59.55	&	\bf59.83*	\\
\hline															
\multirow{4}{*}{\rotatebox[origin=c]{90}{\centering	200K lines}}														
&BLEU	&	33.96	&	33.75	&	33.86	&	32.96	&	33.93	&	33.54	&	33.33	\\
&TER	&	45.64	&	46.17	&	46.14	&	47.04	&	45.99	&	46.26	&	46.77	\\
&METEOR	&	35.01	&	\bf35.08	&	\bf35.08	&	34.96	&	35.01	&	\bf35.09	&	34.98	\\
&CHRF3	&	59.56	&	\bf59.58	&	\bf59.79	&	\bf59.77	&	\bf59.63	&\bf59.85	&	\bf59.65	\\	
\hline															
\multirow{4}{*}{\rotatebox[origin=c]{90}{\centering  500K lines}}															
&BLEU	&	33.75	&	33.72	&	33.74	&	33.13	&	33.75	&	\bf33.95	&	33.46	\\
&TER	&	45.92	&	46.32	&	46.13	&	46.90	&	46.12	&	46.14	&	46.85	\\
&METEOR	&	34.92	&	\bf35.15	&	\bf35.16	&	34.87	&	\bf35.14	&	\bf35.11	&	\bf34.93	\\
&CHRF3	&	59.57	&	59.84	&	59.83	&	59.61	&	\bf59.83	&	\bf60.00*	&	\bf59.77	\\
\hline															
\multicolumn{9}{|c|}{NEWS}\\															
\hline															
\multirow{4}{*}{\rotatebox[origin=c]{90}{\centering  100K lines}}															
&BLEU	&	26.49	&	\bf26.58	&	\bf26.61	&	\bf26.60	&	\bf26.51	&	\bf26.59	&	\bf26.54	\\
&TER	&	54.21	&	54.34	&	54.69	&	54.94	&	54.26	&	54.38	&	54.8	\\
&METEOR	&	30.21	&	\bf30.36*	&	\bf30.43*	&	\bf30.51*	&	\bf30.33*	&	\bf30.49*	&	\bf30.49*	\\
&CHRF3	&	51.80	&	\bf52.09*	&	\bf52.37*	&	\bf52.52*	&	\bf52.06*	&	\bf52.41*	&	\bf52.49*	\\
\hline															
\multirow{4}{*}{\rotatebox[origin=c]{90}{\centering	200K lines}}														
&BLEU	&	26.55	&	\bf26.62	&	\bf26.66	&	\bf26.65	&	\bf26.77*	&	\bf26.72	&	\bf26.80*	\\
&TER	&	54.17	&	54.36	&	54.52	&	54.70	&	54.29	&	54.48	&	54.55	\\
&METEOR	&	30.24	&	\bf30.37*	&	\bf30.49*	&	\bf30.55*	&	\bf30.45*	&	\bf30.54*	&	\bf30.61*	\\
&CHRF3	&	51.89	&	\bf52.14*	&	\bf52.49*	&	\bf52.55*	&	\bf52.35*	&	\bf52.56*	&	\bf52.75*	\\
\hline															
\multirow{4}{*}{\rotatebox[origin=c]{90}{\centering	500K lines}}														
&BLEU	&	26.40	&	\bf26.70*	&	\bf26.92*	&	\bf26.73	&	\bf26.68*	&	\bf26.94*	&	\bf26.98*	\\
&TER	&	54.47	&	\bf54.29	&	\bf54.37	&	\bf54.63	&	\bf54.13*	&	54.35	&	54.56	\\
&METEOR	&	30.1	&	\bf30.44*	&	\bf30.59*	&	\bf30.59*	&	\bf30.42*	&	\bf30.58*	&	\bf30.68*	\\
&CHRF3	&	51.71	&	\bf52.29*	&	\bf52.57*	&	\bf52.66*	&	\bf52.19*	&	\bf52.58*	&   \bf52.81*	\\	
\hline
\end{tabular}
\caption{Results of the models built with different sizes of FDA-selected hybrid data following the \textit{batch} and  \textit{online} approaches. The results in bold indicate an improvement over FDA. The asterisk means the improvement is statistically significant at p=0.01. \label{table:results_FDA_batch_online}}
\end{center}
\end{table*}

In Table \ref{table:results_INR_batch_online} and Table \ref{table:results_FDA_batch_online} we present the results of the models when fine-tuned with a combination of authentic and synthetic data following the \textit{Combine Before Selection} approaches in Section \ref{ssec:hybrid_data_creation}. The tables are structured in two subtables showing the results of \textit{batch} and \textit{online} approaches. Each subtable present the results of three values of $\gamma$: 0.75, 0.50 and 0.25.

In these tables, we see that the performance of the models following the \textit{batch} and \textit{online} approaches is similar. These results are also in accord with those obtained following the \textit{hybr} approach, as the improvements depend more on the domain (most evaluation scores in the NEWS test set indicate statistically significant improvements whereas for the  BIO test set most of them are not) than the TA used, or the value of $\gamma$. Although the best scores tend to be when $\gamma=0.50$ this is not always the case, and moreover we can find experiments in which using high amounts of synthetic sentences (i.e.  $\gamma=0.25$) achieve better results than using a higher proportion of authentic sentences. For instance, in BIO subtable of Table \ref{table:results_INR_batch_online}, using 100K sentences with the \textit{online} $\gamma=0.25$ approach,  the improvements are statistically significant for two evaluation metrics whereas in the other experiments in that row they are not.


When analyzing the translations produced by these models we find several examples in which the translations of models fine-tuned with hybrid data are superior to those tuned with authentic sentences. An example of this is the sentence in the NEWS test set \say{nach Krankenhausangaben wurde ein Polizist verletzt.} (in the reference, \say{according to statements released by the hospital, a police officer was injured.})

This sentence is translated by INR and FDA models (those fine-tuned with 100K authentic sentences) as \say{a policeman was injured after hospital information.}. We see that these models translate the word \say{nach} with its literal meaning (\say{after}) whereas in this context (\say{nach Krankenhausangaben}) it should have been translated as \say{according to} as stated in the reference.

In the hybrid models, we see that the same sentence has been translated as \say{according to hospital information, a policeman was injured.} (in this case the models fine-tuned with hybrid data have produced the same translations). The models tuned with hybrid data are capable of producing the {\em n}-gram \say{according to} which is the same as the reference.

In the selected data, the only sentence containing the {\em n}-gram \say{nach Krankenhausangaben} is the authentic sentence presented in the first row of Table~\ref{table:examples_backtranslated_sentences} (selected by every execution of TA). As we see, this is a noisy sentence as the target-side does not correspond to an accurate translation (observe that in the source sentence we cannot find names such as \say{La Passione} or \say{Carlo Mazzacurati} that are present in the English side). Accordingly, using this sentence in the training of the NMT is harmful.

\begin{table*}[!ht]
\centering
\begin{center}
\begin{tabular}{ |p{0.15cm}|p{4.6cm}|p{4.6cm}|p{4.6cm}|}
\hline
	&	German (auth)	&	 German (synth)	&	English \\
\hline
1	&	\small	nach Krankenhausangaben wurden zwei um die 50 Jahre alte Männer durch das Beben schwer verletzt: einer sei von einem herabfallenden Schornstein getroffen worden, der andere habe durch Glas Schnittwunden erlitten. außerdem seien mehrere Menschen durch herabstürzende Gegenstände in ihren Wohnungen leicht verletzt worden.	&	\small	am Samstag wird es eine weitere Komödie, "La pasone" von Carlo Mazzacurati Italiens, geben, die die fruchtbare Silvio Orlando, die ein washed-up filmin der Toskana ist, in einer Nachbarkapelle aus dem 16. Jh.	&	\small	Saturday will feature another comedy, "La Passione" by Carlo Mazzacurati of Italy starring the prolific Silvio Orlando, who plays a washed-up filmmaker who is forced to set his last-chance project in Tuscany after a plumbing disaster at his country home damages a 16th-century fresco in a neighbouring chapel.	\\
\hline
2	&	\small	die Veranstalter haben viele Konzerte und Recitale geplant. Es wird für uns eine vorzügliche Gelegenheit sein Ihre Freizeit angenehm zu gestalten und Sie für die ernste Musik zu gewinnen.	&	\small	jeder Teilnehmer wird mindestens ein Programm spielen.	&	\small	every participant will play at least one programme.	\\
\hline
3	&	\small	folglich übernimmt Informatic SA keine Gewährleistung für ihre Richtigkeit , ausser sie wurden vom Kunden schriftlich oder per E-Mail ausdrücklich für obligatorisch erklärt .	&	\small. . , . . . . . . . . . . . . . . . . . . . . . . . . . . . . . . . . . . . . . . . . . . .		&	\small	par conséquent, Informatic SA ne donne donc aucune assurance quant à leur exactitude à moins qu'elles n'aient été expressément déclarées obligatoires par écrit ou par e-mail par le client .	\\

\hline
4	&	\small	die Preise liegen zwischen 32.000 und 110.000 Won.	&	\small	der erste Abend beginnt mit einer großen Parade aller Teilnehmer durch die Stadt zum Strand.	&	\small	the first evening starts with a big parade of all participants through the city towards the beach.	\\
\hline
\end{tabular}
\caption{Examples of back-translated sentences}
\label{table:examples_backtranslated_sentences}
\end{center}
\end{table*}






\subsection{Analysis of Back-translated Sentences}
\label{ssec:analysis_back_translated_sentences}

We find many cases where artificially-generated data is more useful for NMT models than authentic translations. In Table \ref{table:examples_backtranslated_sentences} we show some examples.

In rows 1 and 2 we present sentences in which the artificial sentence (\textit{German (synth)} column) is a better translation than the authentic counterpart. In addition to the example described previously (the example of the first row), we also see in row 2 that the authentic candidate pair is (\say{die Veranstalter haben viele Konzerte und Recitale geplant. Es wird für uns eine vorzügliche Gelegenheit sein Ihre Freizeit angenehm zu gestalten und Sie für die ernste Musik zu gewinnen.},\say{every participant will play at least one programme.}) whereas the synthetic counterpart is the pair (\say{jeder Teilnehmer wird mindestens ein Programm spielen.},\say{every participant will play at least one programme.}). In this case, it is preferable to use the synthetic sentence for training instead of the authentic as it is a more accurate translation (observe that the authentic German side consists of two sentences and it is longer than the English-side).

We also present a case in which both authentic and artificial sentences are not proper translations of the English sentence, so both sentences would hurt the performance of NMT if used for training. In row 3 there is a noisy sentence that should not have been included as the target side is not English but French. The TAs search for {\em n}-grams in the source side, so as in this case the artificial sentence consists of a sequence of dots (the BT model has not been able to translate the French sentence) this prevents the TA from selecting it, whereas the authentic sentence-pair could be selected as it is a natural German sentence.





Surprisingly, this correction of inaccurate translations can also be seen on the set of sentences that have been used for training the BT model. As this model does not overfit, when it is provided with the same target sentence used for training, it is capable of generating different valid translations. For example, in row 4 of Table \ref{table:examples_backtranslated_sentences} we see the pair (\say{die Preise liegen zwischen 32.000 und 110.000 Won.},\say{the first evening starts with a big parade of all participants through the city towards the beach.}) which is one of the sentence pairs used for training the BT model. This is a noisy sentence (see, for instance, that the English-side does not include the numbers). However, the sentence generated by the BT model is \say{der erste Abend beginnt mit einer großen Parade aller Teilnehmer durch die Stadt zum Strand.} which is a more accurate translation of than the sentence used for training the model that generates it.

\section{Conclusion and Future Work}




In this work, we have presented how artificially generated sentences can be used to augment a set of candidate sentences so data-selection algorithms have a wider variety of sentences to select from. The TA-selected sets have been evaluated according to how useful they are for improving NMT models.

We have presented three methods of creating such hybrid data: (i) by allowing the TA decide whether to select authentic or synthetic data (\textit{hybr}); (ii) by performing independent executions of the TA on authentic and synthetic sets (\textit{batch}); and (iii) using an MT-generated seed to select monolingual sentences so only the extracted subset is back-translated (\textit{online}).

The experiments showed that artificially-generated sentences can be as competitive as authentic data, as models built with different proportions of authentic and synthetic data achieve similar or even better performance than those fine-tuned with authentic pairs only. On one hand, those sentences whose target-sides could hurt the performance of NMT (such as sentences in a different language to that expected) causes the back-translated sentence to also contain unnatural {\em n}-grams and so TAs would not select them. On the other hand, if the source-side sentence is not an accurate translation of the target side (the problem of comparable corpus), the back-translated counterpart can be a better alternative to use as training data.

In the future, we want to explore other language pairs and other transductive algorithms. Another limitation of this work is that we have augmented the candidate pool with synthetic sentences generated by a single model. We propose to explore whether using several models for generating the synthetic sentences (including different approaches such as combining statistical and neural model \citep{poncelas2019combining}) to augment the candidate pool can cause the selected data to further improve NMT models.

\section*{Acknowledgements}
This research has been supported by the ADAPT Centre for Digital Content Technology which is funded under the SFI Research Centres Programme (Grant 13/RC/2106).

\bibliography{bibl}
\bibliographystyle{acl_natbib}

\end{document}